\documentclass[sigconf]{acmart}
\usepackage{booktabs} 
\usepackage{adjustbox}

\usepackage{bm}
\usepackage{siunitx}
\usepackage{array}
\usepackage{multirow}
\usepackage{subcaption}
\usepackage{colortbl}
\usepackage{xcolor}

\usepackage[justification=justified,skip=2pt]{caption}

\renewcommand{\arraystretch}{0.8}

\makeatletter
\g@addto@macro\normalsize{%
  \abovedisplayskip 3pt plus 2pt minus 3pt%
  \belowdisplayskip \abovedisplayskip
  \abovedisplayshortskip 3pt plus2pt  minus3pt%
  \belowdisplayshortskip 3pt plus2pt minus3pt%
}

\makeatother

\AtBeginDocument{%
  \providecommand\BibTeX{{%
    \normalfont B\kern-0.5em{\scshape i\kern-0.25em b}\kern-0.8em\TeX}}}

\copyrightyear{2024}
\acmYear{2024}
\setcopyright{rightsretained}
\acmConference[WWW '24 Companion]{Companion Proceedings of the ACM Web Conference 2024}{May 13--17, 2024}{Singapore, Singapore}
\acmBooktitle{Companion Proceedings of the ACM Web Conference 2024 (WWW '24 Companion), May 13--17, 2024, Singapore, Singapore}\acmDOI{10.1145/3589335.3651578}
\acmISBN{979-8-4007-0172-6/24/05}

\begin{document}

\title{Unlink to Unlearn: Simplifying Edge Unlearning in GNNs}

\author{Jiajun Tan}
\affiliation{%
  \institution{Institute of Computing Technology, \\ Chinese Academy of Sciences}
  \institution{University of Chinese Academy of Sciences, Beijing, China}
  \city{}
  \country{}
}
\email{tanjiajun22s@ict.ac.cn}

\author{Fei Sun}
\authornote{Corresponding authors.}
\affiliation{%
  \institution{Institute of Computing Technology, \\ Chinese Academy of Sciences}
  \city{Beijing}
  \country{China}
}
\email{sunfei@ict.ac.cn}

\author{Ruichen Qiu}
\affiliation{%
  \institution{University of Chinese Academy of Sciences, Beijing, China}
  \city{}
  \country{}
}
\email{quiruichen20@mails.ucas.ac.cn}

\author{Du Su}
\affiliation{%
  \institution{Institute of Computing Technology, \\ Chinese Academy of Sciences}
  \city{Beijing}
  \country{China}
}
\email{sudu@ict.ac.cn}

\author{Huawei Shen}
\authornotemark[1]
\affiliation{%
  \institution{Institute of Computing Technology, \\ Chinese Academy of Sciences}
  \city{Beijing}
  \country{China}
}
\email{shenhuawei@ict.ac.cn}

\renewcommand{\authors}{Jiajun Tan, Fei Sun, Ruichen Qiu, Du Su, Huawei Shen}
\renewcommand{\shortauthors}{Jiajun Tan, Fei Sun, et al.}

\begin{abstract}

As concerns over data privacy intensify, unlearning in Graph Neural Networks (GNNs) has emerged as a prominent research frontier in academia.
This concept is pivotal in enforcing the \textit{right to be forgotten}, which entails the selective removal of specific data from trained GNNs upon user request.
Our research focuses on edge unlearning, a process of particular relevance to real-world applications. 
Current state-of-the-art approaches like GNNDelete can eliminate the influence of specific edges yet suffer from \textit{over-forgetting}, which means the unlearning process inadvertently removes excessive information beyond needed, leading to a significant performance decline for remaining edges.
Our analysis identifies the loss functions of GNNDelete as the primary source of over-forgetting and also suggests that loss functions may be redundant for effective edge unlearning.
Building on these insights, we simplify GNNDelete to develop \textbf{Unlink to Unlearn} (UtU), a novel method that facilitates unlearning exclusively through unlinking the forget edges from graph structure.
Our extensive experiments demonstrate that UtU delivers privacy protection on par with that of a retrained model while preserving high accuracy in downstream tasks, by upholding over 97.3\% of the retrained model's privacy protection capabilities and 99.8\% of its link prediction accuracy. 
Meanwhile, UtU requires only constant computational demands, underscoring its advantage as a highly lightweight and practical edge unlearning solution.

\end{abstract}

%



\maketitle

\section{Introduction}

Although Graph Neural Networks (GNNs) have achieved great success in various tasks~\cite{wu2021survey}, this advancement inherently comes with the risk of privacy leakage, as training data with sensitive personal information can be implicitly ``remembered'' within model parameters.
To address these concerns, recent legislations, e.g., EU’s General Data Protection Regulation (GDPR) and the California Consumer Privacy Act (CCPA), have granted individuals with the \textit{right to be forgotten}, enabling them to request elimination of their private data from online platforms. 
Consequently, machine unlearning~\cite{cao2015towards} has emerged, allowing quick and efficient removal of specific data from a trained model, rather than retraining from scratch. 

In this paper, we focus on edge unlearning, a key unlearning scheme in graphs, owing to its pivotal role in real-world applications such as safeguarding edge privacy in social networks.
Consider the scenario where individuals in online social networks may seek to conceal certain private social connections.
In these instances, GNNs that have been trained on these graphs require timely updates to eliminate any influence of the data intended to be forgotten, while preserving performance on retrained edges.

Recently, GNNDelete has achieved state-of-the-art performance in edge unlearning, however, we observed a considerable decline in its prediction accuracy for edges in the retained training set, especially for those resembling or closely associated with the edges subjected to unlearning.
We introduce the term \textit{over-forgetting} to describe such a phenomenon, where an unlearning algorithm inadvertently eliminates an excessive amount of information from the retained data.
In this context, ``excessive'' refers to a scenario where the performance of the unlearned model on retained data deteriorates significantly compared to a model retrained from scratch using only the retained data.

In this study, we address the challenge of over-forgetting by introducing \underline{\textbf{U}}nlink \underline{\textbf{t}}o \underline{\textbf{U}}nlearn (\textbf{UtU}). 
Our investigation has revealed deficiencies in the design of GNNDelete's loss functions. One loss function designed to eliminate the influence of forget edges opts for an unsuitable optimization objective, being the primary contributor to over-forgetting. 
The other, designed to alleviate the issue of over-forgetting, fails to prevent the performance decline of retaining edges as intended. 
In light of these findings, we deprecate both loss functions in GNNDelete, while facilitating edge unlearning by unlinking forget edges from the original graph structure. 
Our method eliminates the need for complex parameter optimization, reducing computation overhead by orders of magnitude. 
Our experimental evaluations indicate that UtU's performance on downstream tasks, its efficacy in unlearning, and its output distribution are more aligned with those of the retrained model, which is broadly regarded as the gold standard of unlearning.

\section{Preliminaries}

\subsection{Operation of GNNs}
\label{para:lp} 
Consider a graph $G=(V, E)$ with node set $V$ and edge set $E$. 
Each node $v_i \in V$ is often associated with a feature vector $\bm{x}_i$. 
A GNN model $\mathcal{M}$, parameterized by $\boldsymbol{\theta}$, is composed of multiple GNN layers, which process node features and graph structural information to generate node embeddings via the message passing mechanism. 

Initialized by $\bm{h}_i^{0} = \bm{x}_i$ for each node $v_i$, the operation of $l$-th GNN layer can be formally expressed as follows:
\begin{align}
    \bm{m}_i^{l} &= \mathsf{msg}(\bm{h}_i^{l-1},\, \{ \bm{h}_j^{l-1} \mid j \in \mathcal{N}(i) \}), \label{eq:1} \\ 
    \bm{h}_i^{l} &= \mathsf{upd} (\bm{h}_i^{l-1},\, \bm{m}_i^{l}) \label{eq:2}
\end{align}
where $\mathcal{N}(i)$ denotes neighborhood of $v_i$, and $\mathsf{msg}(\cdot)$ and $\mathsf{upd}(\cdot)$ vary among different GNN types, representing message function and update function. The final output of $\mathcal{M}$ is the node embeddings of last layer, denoted by $\bm{h} = \mathcal{M}(G, \theta)$. 

\subsection{Edge Unlearning on GNNs}

Let $\mathcal{M}_0$ be a randomly initialized GNN, and let the original model $\mathcal{M}^*$ be trained using a learning algorithm $\mathcal{A}(\mathcal{M}_0, G)$ on graph $G$. 
The forget set, defined as $E_d \subseteq E$, contains the edges requested for removal, while the retain set $E_r = E \backslash E_d$ represents the remaining edges in the training graph $G$. 
The objective of unlearning is to devise an unlearning process $\mathcal{U}$ that renders the unlearned model $\mathcal{M}_u = \mathcal{U}(\mathcal{M}^*, G, E_d)$ indistinguishable from the retrained model $\mathcal{M}_r = \mathcal{A}(\mathcal{M}_0, G_r)$, with $G_r=(V,E_r)$ being the retain graph.

A variety of unlearning algorithms have been developed to address edge unlearning requests in GNNs.
\cite{Chen:RecUn,chen2022graph} leverages SISA paradigm~\cite{bourtoule2021mu}, where only the sub-model corresponding to the removed data needs to be retrained and then aggregates its result with other sub-models. 
Other strategies directly update the original model, including leveraging the influence function~\cite{wu2023gif}, providing a theoretical guarantee of unlearning via differential privacy~\cite{wu2023certified}, projecting parameters to irrelevant subspace~\cite{cong2023efficiently}.

\subsection{GNNDelete}
Recently, GNNDelete~\cite{cheng2023gnndelete} surpasses various baseline methods, showing a strong capability to unlearn selected edges by a learning-to-unlearn framework.
It first inserts a linear transformation $\phi$ with learnable parameters after each GNN layer, while freezing the original model's parameters. 
$\phi$ is only applied to nodes in $l$-hop enclosing subgraph of forget edge $e_{uv}$, namely $S^l_{uv}$ , by transforming each node's original embedding into unlearned embedding: $\bm{h}'^{l}_i = \phi^l (\bm{h}_i^l)$; for other nodes, their embeddings remain unchanged.

Two loss functions, \textit{Deleted Edge Consistency} (DEC) loss $\mathcal{L}_{DEC}$ and \textit{Neighborhood Influence} (NI) loss $\mathcal{L}_{NI}$, are then computed layer-wise. 
Generally, DEC loss is intended for unlearning edges in $E_d$, while the NI loss aims to repair node embeddings in $S^l_{uv}$. 
We will discuss their design in more detail in \ref{ref:analysis}.
During the backward pass, the parameters of $\phi$ are optimized based on the weighted total loss, which is represented as: 
\begin{equation}
     \mathcal{L}^l = \lambda \mathcal{L}_{DEC}^l + (1 - \lambda)\mathcal{L}_{NI}^l,
\end{equation}
GNNDelete uses $\lambda = 0.5$ to report its performance, claiming this setting can achieve the best overall performance.

\section{UtU: Unlink to Unlearn}

While GNNDelete reports a prominent unlearning ability, it suffers from the issue of over-forgetting. In this section, we first provide a formal introduction to the over-forgetting problem, subsequently revealing the connection between GNNDelete's inappropriate loss and over-forgetting. 
By removing these loss functions, we further introduce our simplified approach \textbf{UtU} to alleviate over-forgetting.

\subsection{Over-forgetting in Edge Unlearning}

\begin{figure}[tbp]
  \centering
  \begin{subfigure}[b]{0.45\linewidth}
    \includegraphics[width=\linewidth]{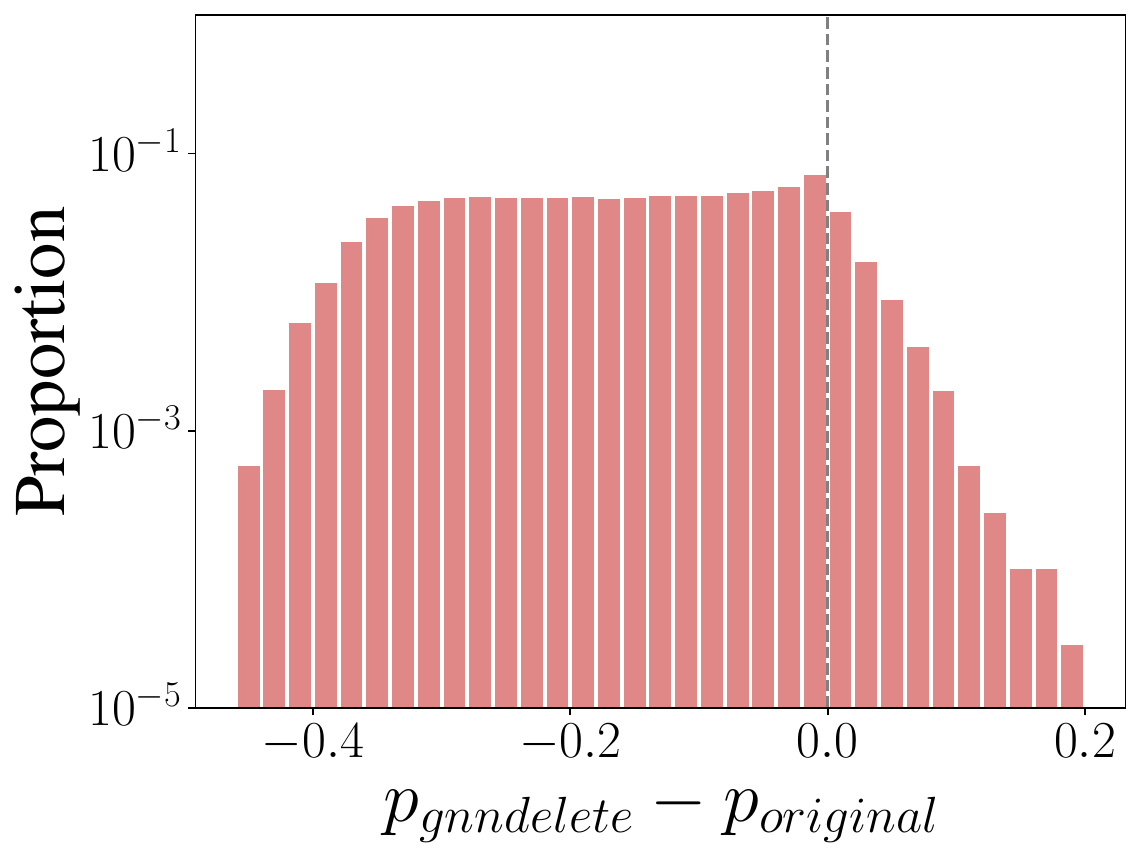}
    \caption{Unlearn by GNNDelete.}
    \label{fig:of-sub1}
  \end{subfigure}
  \hfil 
  \begin{subfigure}[b]{0.45\linewidth}
    \includegraphics[width=\linewidth]{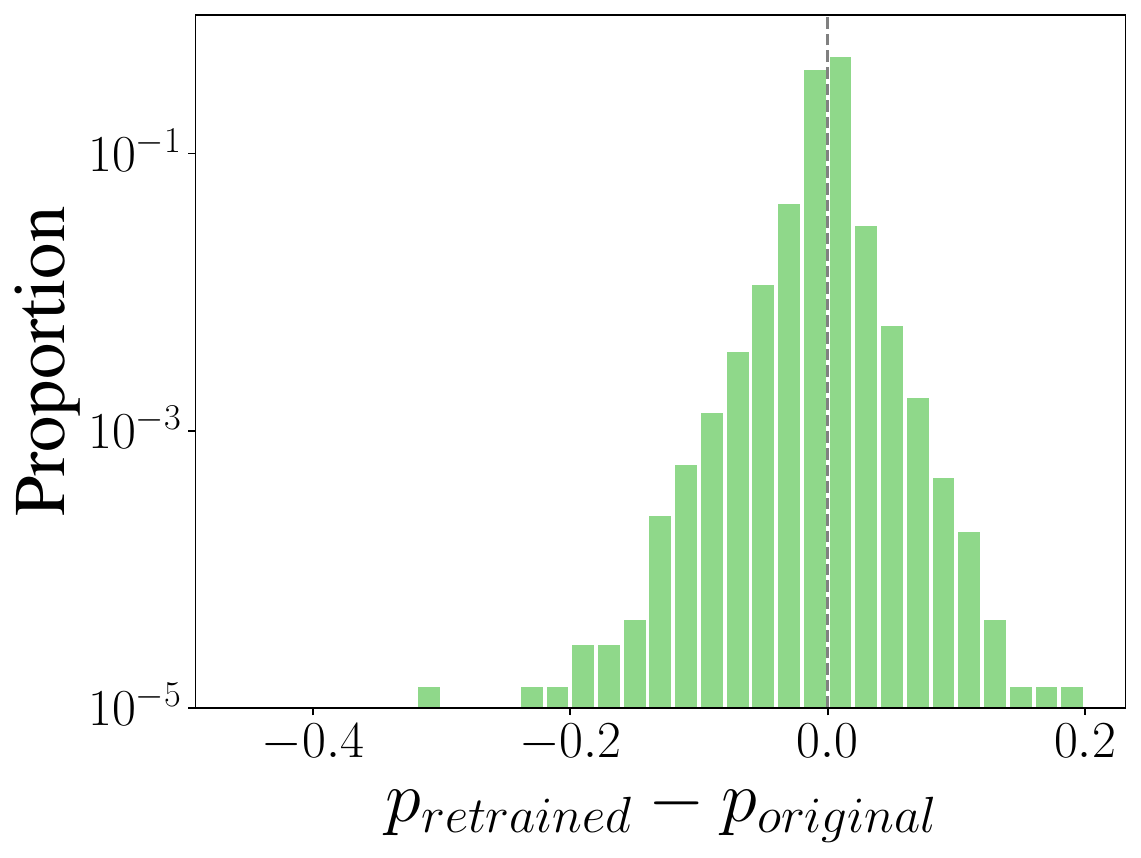}
    \caption{Retrain from scratch.}
    \label{fig:of-sub2}
  \end{subfigure}
  \caption{An intuitive demonstration of over-forgetting, comparing the difference of retaining edge predictions before and after forgetting. The x-axis represents the change in predicted probabilities after unlearning, and the y-axis shows the distribution of these changes across the retain set $E_r$.}
  \label{fig:of}
\end{figure}

As previously introduced, over-forgetting refers to the phenomenon where the performance of retain set significantly deteriorates after unlearning. 
Figure \ref{fig:of} shows the over-forgetting observed on GNNDelete. 
Despite only 5\% of the edges being unlearned, a substantial 92.4\% of the retained edges experience performance decrement. 
Conversely, the retrained model's predictions for the majority of retained edges remain mostly unchanged.
In order to quantitatively assess over-forgetting, we compare the performance of unlearned model against a retrained model to gauge the impact of the forgetting procedure. 

For link prediction task, the probability of the existence of the edge between nodes $(v_i, v_j)$ is predicted by integrating their final embeddings $\bm{h}^L_i, \bm{h}^L_j$ using a score function $\varphi(\cdot)$, as $p_{ij} = \varphi(\bm{h}^L_i, \bm{h}^L_j)$.
\label{ref:deltap}
For $e_{ij} \in E_r$, we identify over-forgetting if the predicted probability of $e_{ij}$ in $\mathcal{M}_u$ decreases compared to $\mathcal{M}_r$, i.e., $\Delta p_{ij} = \varphi(\bm{h}'_i, \bm{h}'_j) - \varphi(\bm{h}^r_i, \bm{h}^r_j) < 0$, where $\bm{h}'$ and $\bm{h}^r$ represent embeddings generated by models $\mathcal{M}_u$ and $\mathcal{M}_r$, respectively. 
Typically, our focus is on the overall performance decline across $E_r$, which can be measured as $\Delta p_r = \mathsf{mean}(\Delta p_{ij}), \forall e_{ij} \in E_r$.

\subsection{Analysis on Unlearning Target}
\label{ref:analysis}
Now we discuss the design of GNNDelete's loss functions to find out the source of over-forgetting. The Deleted Edge Consistency (DEC) loss minimizes the difference between predictions of forget edges $e_{uv}$ and random-chosen node pairs: 
\begin{equation}
    \mathcal{L}_{\mathrm{DEC}}^l = \mathsf{mse}(\{ [\bm{h}'^{l}_u; \bm{h}'^{l}_v] \mid e_{uv} \in E_d \}, \{ [\bm{h}_p^l; \bm{h}_q^l] \mid p, q \in_R V \}), 
\end{equation}
where $[\cdot; \cdot]$ denotes the concatenation of two vectors. $\mathsf{mse}$ refers to Mean-Squared Error. $\in_R$ means randomly chosen. 

However, random node pairs may not be a suitable optimization target. The optimization of DEC loss will introduce structural noise by encouraging node embeddings in $E_d$ to reflect random connections rather than actual graph topology, leading to inaccurate representations and prediction results. Moreover, this noise propagates to neighboring nodes by message passing mechanism as pointed out in Sec.~\ref{para:lp}, degrading the embedding quality on a broader scale and exacerbating the issue of over-forgetting.

Besides, connected nodes in a graph tend to share similar attributes or belong to the same class according to the homophily hypothesis ~\cite{mcpherson2001homophily}. 
Despite being removed due to unlearning requests, the samples in $E_d$ originate from pre-existing edges in the graph, implying that their end nodes ought to exhibit strong homophily, and should not be equated with arbitrarily selected node pairs.

\textit{Neighborhood Influence} (NI) loss base on the idea that removing $e_{uv}$ should not affect the predictions of its enclosing subgraph $S_{uv}$:
\begin{equation}
    \mathcal{L}_{\mathrm{NI}}^l = \mathsf{mse}(\Vert_w \{ [\bm{h}'^l_w] \mid w \in S_{uv}^l / e_{uv}^l\},\, \Vert_w \{[\bm{h}_w^l] \mid w \in S_{uv}^l\}),
\end{equation}
where $\Vert$ signifies the concatenation of multiple vectors. 

NI loss guides the unlearned embedding $\bm{h}'^l_w$ to match the original one, acting as a regularization term for node embeddings in $S^l_{uv}$ to mitigate the structural noise induced by DEC loss. 
However, note that node features and edges are combined as GNN's input, the original embedding is generated using the original graph without removing $e_{uv}$'s influence. That means the unlearned embedding will still contain structural information of edges in forget set, which weakens its ability to repair over-forgetting.

\subsection{UtU: a Minimalist Approach}
Given the preceding analysis, we propose to eliminate the DEC loss due to its selection of an unsuitable target for forgetting.
Following the removal of DEC loss, the NI loss, initially serving as a corrective measure for DEC loss, is also deemed unnecessary.
We then propose that edge unlearning can be effectively achieved by only altering the input edge indexes to that of retain graph $G_r$, which can be named \textbf{Unlink to Unlearn}:
\begin{equation}
    \bm{h}' = \mathcal{M}^*(G_r; \theta^*),
\end{equation}
where $\theta^*$ denotes parameters of original model $\mathcal{M^*}$.

This approach is grounded on the insights of GNN operations. 
The edges in the graph mainly facilitate message passing between node features, as delineated in Eq.~\ref{eq:1}. 
Most GNN models' parameters are located during the node features' update step, as specified in Eq. \ref{eq:2}, which does not involve edge utilization.
Therefore, removing a forgotten edge during inference can effectively block message propagation from neighboring nodes linked to the forget set. 
This action alone suffices to eliminate the edge's influence from the model, thereby achieving our unlearning objective.

It is worth noting that UtU is remarkably efficient through its minimalistic design. Deleting an edge from the graph structure only requires $\mathcal{O}(1)$ time complexity, making UtU a nearly ``zero-cost'' edge unlearning solution.

\section{Experiments}

\subsection{Experimental Setup}

\textbf{Datasets}.
We conduct the experiments on four real-world datasets, including citation networks: CoraFull~\cite{bojchevski2018deep} and PubMed~\cite{bojchevski2018deep}, and collaboration networks: CS~\cite{shchur2018pitfalls} and OGB-collab~\cite{hu2020ogb}. 
Table \ref{dataset_stat} shows details of these datasets. 
In accordance with previous work, we split 90\% edges for the training set, 5\% for validation, and 5\% for test.

\begin{table}[]
\centering
\renewcommand{\arraystretch}{0.85}
\caption{Statistical Overview of the Datasets.}
\label{dataset_stat}
\begin{tabular}{lrrrr}
\toprule
\textbf{Dataset} & \textbf{\# Nodes} & \textbf{\# Edges} & \textbf{\# Features} & \textbf{\# Classes} \\ 
\midrule
CoraFull         & 19,793   & 126,842 & 8,710   & 70   \\ 
PubMed           & 19,717   & 88,648 &  500  & 3   \\ 
CS          & 18,333   & 163,778 &  6,805   & 15      \\ 
OGB-collab & 235,868    & 2,238,104 & 128  & N/A  \\
\bottomrule
\end{tabular}
\end{table}

\textbf{Baselines.} We choose most widely-used 2-layer GNNs as backbones: GCN~\cite{kipf2017semisupervised}, GAT~\cite{veličković2018graph}, and GIN~\cite{xu2018how}. 
We also consider following unlearning methods for comparison: Retrain from scratch, Gradient Ascent, GIF~\cite{wu2023gif},  GNNDelete~\cite{cheng2023gnndelete}, and a variant of GNNDelete by only removing DEC loss, namely GNNDelete-NI.

\textbf{Tasks}. 
Following the common practice in~\cite{cheng2023gnndelete,wu2023certified}, we train all models on link prediction task and then perform edge unlearning. 
Forget edges are randomly chosen from the training set. We vary the proportion of forget edges from 0.1\% to 5\% to examine algorithm performance under different scenarios. There are few scenarios where more than 5\% of edges need to be unlearned simultaneously.

\textbf{Metrics.} We adopt ROC-AUC to evaluate downstream tasks for link prediction. To assess the effectiveness of unlearning, we compare the unlearned model against the retrained model using JS divergence and ROC-AUC of MI Attack. Additionally, we use $\Delta p$, introduced in Sec.~\ref{ref:deltap}, to compare over-forgetting.

\textbf{Implementation.} We follow the default hyper-parameter setting of all baselines, and metrics are reported across an average of five independent runs. Codes are available at provided link\footnote{https://github.com/Sumsky21/Unlink-to-Unlearn}.

\subsection{Experiment Result and Anlysis}
\subsubsection{Downstream Task}

In this part, we compare the utility of unlearned models obtained by different unlearning methods, as we anticipate that unlearning will not harm the performance of GNN on downstream tasks. ROC-AUC was used to determine the model's ability to predict hidden test edges. Results are shown in Table \ref{lp-auc}, where UtU performs the best in most settings, with the closest gap of 0.001 on average compared to retraining. 

\begin{table}[]
    \centering
    \caption{AUC ($\uparrow$) on Link Prediction. Forget Set: 5.0\% edges.}
    \label{lp-auc}
    \renewcommand{\tabcolsep}{4pt}
    \begin{adjustbox}{max width=\linewidth} 
    \begin{tabular}{ll>{\columncolor{gray!20}}cccccc}
    \toprule
    Dataset & Model & Retrain & GradAscent & GIF & GNNDelete & GNNDelete-NI & UtU\\ \midrule
     \multirow{3}{*}{CoraFull}  & GCN & 0.967 & 0.563 & 0.964 & 0.922 & \textbf{0.967} & 0.965\\
     & GAT & 0.963 & 0.766 & 0.926 & 0.934 & 0.947 & \textbf{0.964}\\
     & GIN & 0.961 & 0.596 & 0.742 & 0.897 & 0.958 & \textbf{0.960}\\ \midrule
     \multirow{3}{*}{PubMed} & GCN & 0.970 & 0.375 & 0.924 & 0.934 & 0.968 & \textbf{0.969} \\
     & GAT & 0.936 & 0.766 & 0.774 & 0.890 & 0.927 & \textbf{0.933} \\
     & GIN & 0.939 & 0.545 & 0.842 & 0.887 & 0.938 & \textbf{0.942} \\ \midrule
     \multirow{3}{*}{CS} & GCN & 0.968 & 0.786 & 0.950 & 0.947 & 0.968 & \textbf{0.970} \\
     & GAT & 0.963 & 0.846 & 0.941 & 0.943 & 0.958 & \textbf{0.963} \\
     & GIN & 0.960 & 0.583 & 0.520 & 0.900 & 0.959 & \textbf{0.960} \\ \midrule
     \multirow{3}{*}{OGB-collab} & GCN & 0.985 & 0.406 & 0.971 & 0.925 & 0.981 & \textbf{0.987} \\
     & GAT & 0.971 & 0.755 & 0.744 & 0.924 & 0.960 & \textbf{0.971} \\
     & GIN & 0.925 & 0.683 & 0.500 & 0.805 & 0.912 & \textbf{0.913} \\
    \bottomrule
    \end{tabular}
    \end{adjustbox}
\end{table}

\begin{table}[]
    \centering
    \caption{AUC on MI Attack. Forget Set: 5.0\% edges.}
    \label{mia-auc}
    \renewcommand{\tabcolsep}{4pt}
    \begin{adjustbox}{max width=\linewidth} 
    \begin{tabular}{ll>{\columncolor{gray!20}}cccccc}
    \toprule
    Dataset & Model & Retrain & GradAscent & GIF & GNNDelete & GNNDelete-NI & UtU\\ \midrule
     \multirow{3}{*}{CoraFull}  & GCN & 0.580 & 0.500 & 0.529 & 0.712 & 0.531 & 0.528\\
     & GAT & 0.582 & 0.500 & 0.547 & 0.719 & 0.543 & 0.541\\
     & GIN & 0.586 & 0.510 & 0.508 & 0.721 & 0.586 & 0.605\\ \midrule
     \multirow{3}{*}{PubMed} & GCN & 0.616 & 0.500 & 0.555 & 0.652 & 0.555 & 0.551 \\
     & GAT & 0.624 & 0.500 & 0.574 & 0.708 & 0.576 & 0.578 \\
     & GIN & 0.603 & 0.519 & 0.552 & 0.752 & 0.662 & 0.625 \\ \midrule
     \multirow{3}{*}{CS} & GCN & 0.593 & 0.500 & 0.580 & 0.566 & 0.573 & 0.577 \\
     & GAT & 0.574 & 0.500 & 0.543 & 0.615 & 0.540 & 0.542 \\
     & GIN & 0.580 & 0.653 & 0.497 & 0.666 & 0.592 & 0.590 \\ \midrule
     \multirow{3}{*}{OGB-collab} & GCN & 0.515 & 0.500 & 0.538 & 0.542 & 0.528 & 0.547 \\
     & GAT & 0.560 & 0.500 & 0.533 & 0.484 & 0.478 & 0.554 \\
     & GIN & 0.571 & 0.502 & 0.500 & 0.511 & 0.555 & 0.556 \\ \midrule
     \multicolumn{3}{c|}{Avg Diff. with Retrain (\%)} & 6.67 & 4.40 & 5.53 & 2.21 & \textbf{1.58} \\
    \bottomrule
    \end{tabular}
    \end{adjustbox}
\end{table}

\subsubsection{Unlearning Efficacy}

\begin{figure}[]
  \centering
  \includegraphics[width=\linewidth]{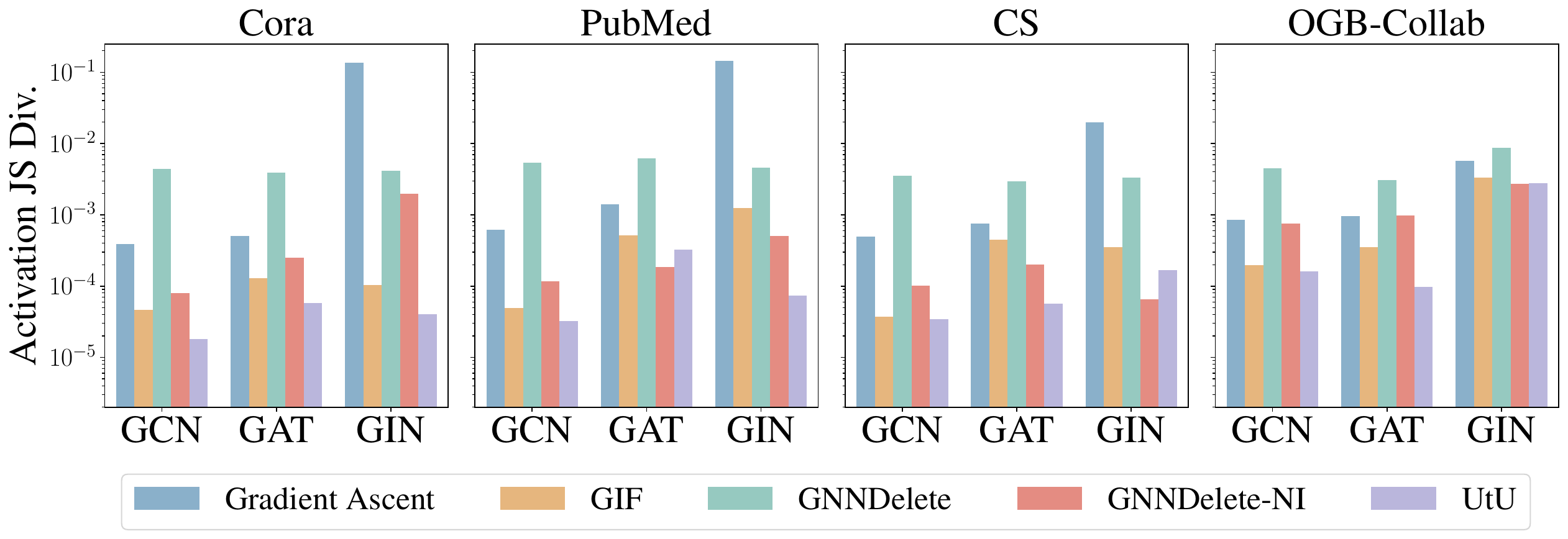}
  \caption{Activation distance ($\downarrow$) on forget set (5.0\% edges). }
  \label{fig:jsd}
\end{figure}

The model after unlearning should treat the forget edges as if it had never seen them before. 
Hence, we expect predictions on these edges to be similar to those of a model that has been retrained from scratch.
Following~\cite{wu2021survey}, we use the activation distance measured by JS divergence, along with membership inference (MI) attack~\cite{olatunji2021membership} to assess whether the model has truly achieved the effect of unlearning. 
Figure \ref{fig:jsd} illustrates that the outputs of forget edges from UtU closely mirror those from the retrained model. Table \ref{mia-auc} shows that the resistance to MI Attack of UtU is generally more aligned with retraining than baseline methods.

\subsubsection{Over-forgetting evaluation}

\begin{figure}[]
  \centering
  \includegraphics[width=0.9\linewidth]{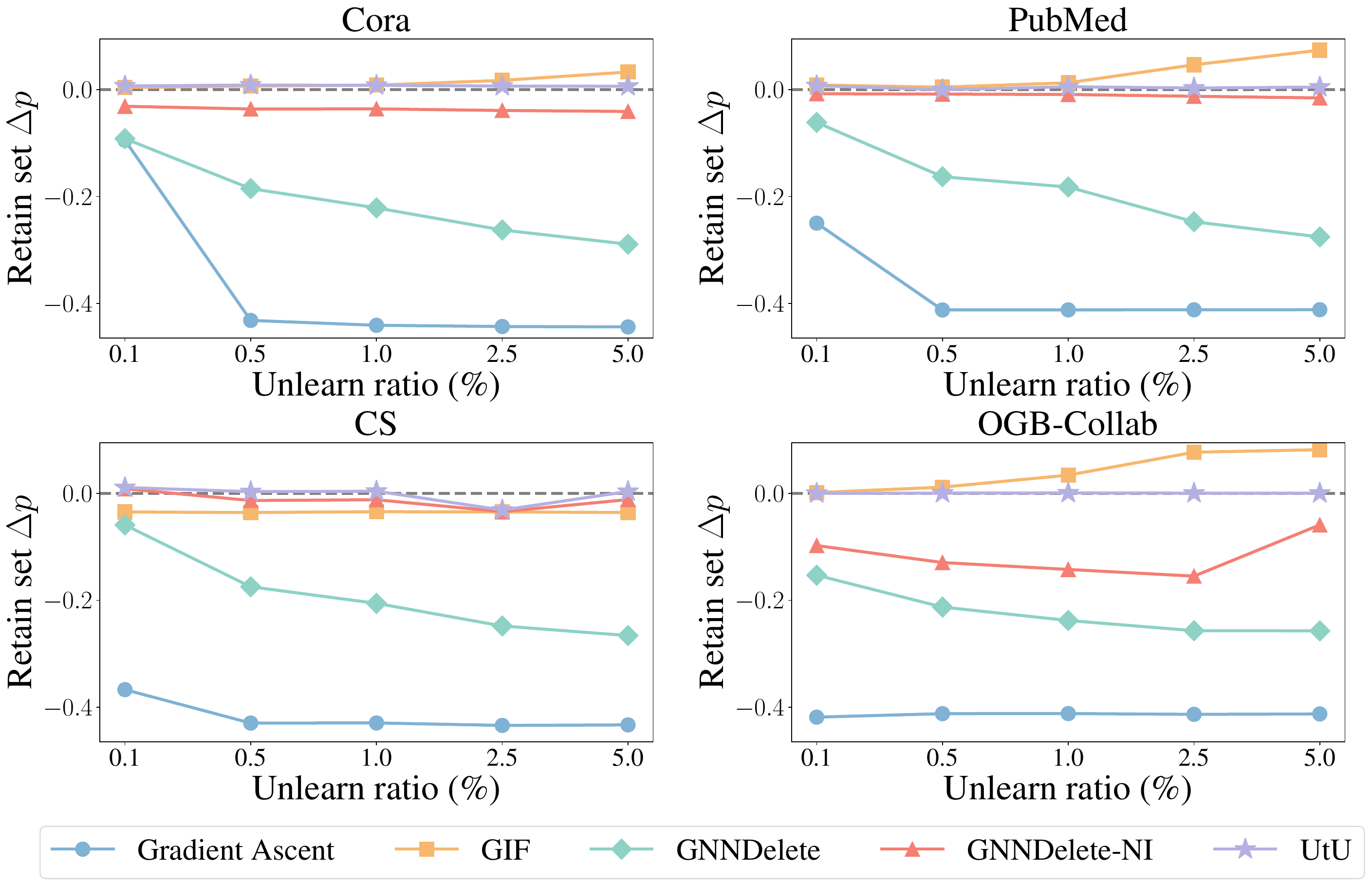}
  \caption{Comparison of over-forgetting on GAT backbone.}
  \label{fig:dp}
\end{figure}

Figure \ref{fig:dp} shows the trend of $\Delta p$ as the size of the forget set changes. 
As described in \ref{ref:deltap}, $\Delta p$ represents the average difference of the edge predictions of the retain set, compared with that of retrained. 
Lower $\Delta p$ indicates more serious over-forgetting. 
The results indicate that our method remains unaffected by over-forgetting, regardless of the forget set's size. 
Furthermore, its predictions of retain set are also highly consistent with those of the retrained model.

\section{Conclusion}

In this work, we address the issue of over-forgetting in the state-of-the-art edge unlearning method, GNNDelete.
Our analysis identifies a correlation between its loss functions and the over-forgetting problem. To mitigate this, we introduce a simplified approach named Unlink to Unlearn (UtU).
UtU effectively eliminates the influence of forgotten edges by merely unlinking the forgotten edges, thus obstructing the corresponding message-passing paths in GNN during the inference stage. 
Experimental results demonstrate that UtU acts on par with the retrained model with near-zero computational overhead. 
Our findings suggest that removing a small number of edges might have little influence on the model parameters, highlighting an avenue for future research to investigate this phenomenon.

\begin{acks}
This work was supported by the National Key R\&D Program of China (2022YFB3103700, 2022YFB3103704), the National Natural Science Foundation of China under Grant No.~U21B2046, and the Innovation Funding of ICT, CAS under Grant No.~E361120. 
\end{acks}

\bibliographystyle{ACM-Reference-Format}
\balance
\bibliography{reference.bib}

\appendix

\end{document}